\documentclass[10pt,twocolumn,letterpaper]{article}

\usepackage{cvpr}
\usepackage{times}
\usepackage{epsfig}
\usepackage{graphicx}
\usepackage{amsmath}
\usepackage{amssymb}

\usepackage[pagebackref=true,breaklinks=true,letterpaper=true,colorlinks,bookmarks=false]{hyperref}

\cvprfinalcopy 

\def\cvprPaperID{1240} 

\ifcvprfinal\fi
\begin{document}

\title{Pooling the Convolutional Layers in Deep ConvNets for Action Recognition}

\author{Shichao Zhao$^1$, Yanbin Liu$^1$, Yahong Han$^1$, Richang Hong$^2$\\
$^1$School of Computer Science and Technology, Tianjin University, China\\
$^2$School of Computer and Information, Hefei University of Technology, China\\
{\tt\small \{zhaoshichao, csyanbin, yahong\}@tju.edu.cn, hongrc.hfut@gmail.com}
}

\maketitle

\begin{abstract}
   Deep ConvNets have shown its good performance in image classification tasks. However it still remains as a problem in deep video representation for action recognition. The problem comes from two aspects: on one hand, current video ConvNets are relatively shallow compared with image ConvNets, which limits its capability of capturing the complex video action information; on the other hand, temporal information of videos is not properly utilized to pool and encode the video sequences.

   Towards these issues, in this paper, we utilize two state-of-the-art ConvNets, i.e., the very deep spatial net (VGGNet \cite{VGG}) and the temporal net from Two-Stream ConvNets \cite{two-stream}, for action representation. The convolutional layers and the proposed new layer, called frame-diff layer, are extracted and pooled with two temporal pooling strategy: Trajectory pooling and line pooling. The pooled local descriptors are then encoded with VLAD to form the video representations. In order to verify the effectiveness of the proposed framework, we conduct experiments on UCF101 and HMDB51 datasets. It achieves the accuracy of 93.78\% on UCF101 which is the state-of-the-art and the accuracy of 65.62\% on HMDB51 which is comparable to the state-of-the-art.

\end{abstract}

\section{Introduction}
Human action recognition \cite{iDT} \cite{human-action1} \cite{DT} \cite{UCFiDT+FV} has attracted much attention in recent years due to its potential applications in automatic video analysis, video surveillance, sports event analysis and virtual reality etc. Still image classification \cite{AlexNet} has gained great success in recent years, whereas human action recognition remains as a problem especially in realistic videos like movies, sports videos and daily-life consumer videos. The problem is caused by some inherent characteristics of action videos such as intra-class variation, occlusions, view point changes, background noises, motion speed and actor differences.

In early researches of action recognition, people have designed effective hand-crafted descriptors (Stip \cite{stip}, MoSift \cite{mosift}, DT \cite{DT}, iDT \cite{iDT}) to capture the spatial and temporal information from video sequences for action recognition. Among these descriptors, improved Dense Trajectories (iDT) has dominated video analysis owing to its good performance. Despite the good performance, iDT has its weakness in huge computation costs \cite{LCD} and large disk affords \cite{sapienza2014feature}. For example, it takes 160GB and 679GB memory respectively to store the extracted features of HMDB51 \cite{HMDB51} and UCF101 \cite{UCF101} datasets.

Due to the constrains of hand-crafted features and the success of deeply learned features of images, researchers have tried to generate video representations from deep ConvNets. A series of attempts like 3D-CNN \cite{3D-CNN}, Deep ConvNets \cite{DeepConvNets}, Two-Stream ConvNets \cite{two-stream} have been proposed. However, unlike image classification \cite{AlexNet}, deep video ConvNets did not make great progress over traditional local descriptors like iDT. We find that there are mainly two reasons that hinder the performance of deep video representations.

\begin{table}
\begin{center}
\begin{tabular}{|l|c|l|l|l|}
\hline
Team & Year & Place & Error(top-5) & Depth\\
\hline\hline
SuperVision & 2012 & 1st & 15.3\% & 8 \\
Clarifai    & 2013 & 1st & 11.7\% & 8 \\
MSRA        & 2014 & 3rd & 7.35\% & 8 \\
VGG         & 2014 & 2nd & 7.32\% & 16-19\\
GooleLeNet  & 2014 & 1st & 6.67\% & 22\\
\hline
\end{tabular}
\end{center}
\caption{Recent ImageNet classification results \cite{szegedy2015going}. We can see that deeper ConvNets usually obtain better performance with proper training.}
\label{depth}
\end{table}

\begin{figure*}[t]
\begin{center}
\epsfig{file=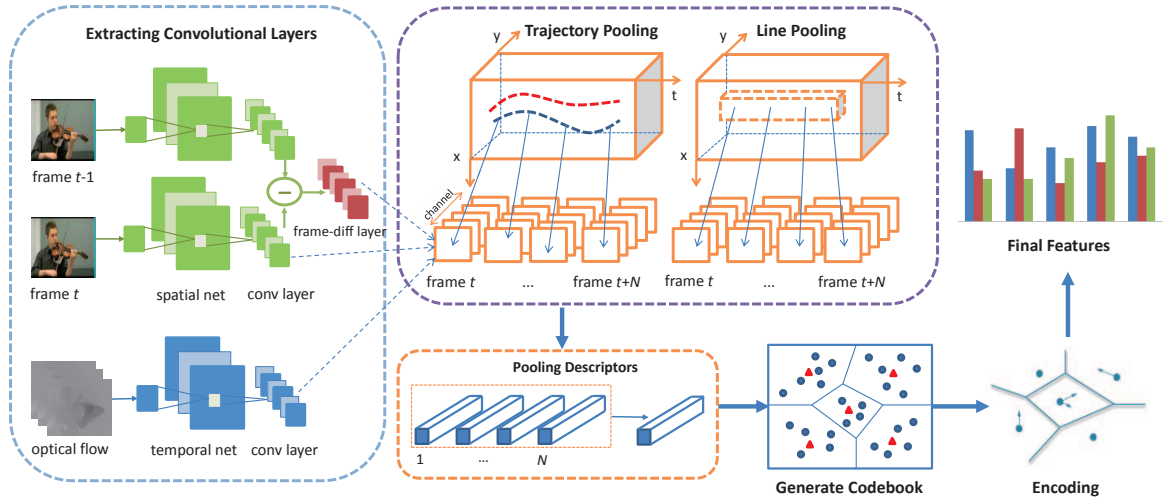, width=0.9\textwidth}
\end{center}
   \caption{The proposed convolutional layers pooling framework. At first, we choose the very deep VGGNet \cite{VGG} as the spatial net and optical flow nets from Two-Stream ConvNets \cite{two-stream} as the temporal nets. Then we extract feature maps from convolutional layer and frame-diff layer in the spatial nets and the convolutional layers in temporal flow nets. The feature maps are pooled by two strategies: Trajectory pooling and line pooling. Finally, the pooled features are encoded by VLAD through codebook generation and quantization steps to get final representations. }
\label{fig:big_pic}
\end{figure*}

Firstly, current video ConvNets is relatively shallow compared with image ConvNets, which limits its capability of capturing the complex video action information. In image, it only contains 2D spatial information like low-level color, texture features and high-level object concepts. While in video, it contains more complex information like scene context, interacting objects, human pose and motion speed. In our common sense, the more complex problem usually need more effective and powerful model to deal with. From Table \ref{depth} we can see that in the large-scale image tasks like ImageNet classification, deeper and more effective models can usually achieve higher performances. However, the representation capacity of current video ConvNets is constrained by their depth. For example, 3D-CNN \cite{3D-CNN} only contains 4 weighted layers (3 convolutional and 1 fully-connected layers), while Deep ConvNets \cite{DeepConvNets} and Two-Stream ConvNets contain 8 layers (5 convolutional and 3 fully-connected layers). As to image, VGG contains 16-19 layers (5 groups of convolutional and 3 fully-connected layers) and GoogleNet contains 22 layers (9 Inception modules), which is much deeper and more informative.

Secondly, temporal information of videos is not properly utilized to pool and encode the video sequences. Pooling and coding are two key factors in both hand-crafted and deep feature representations for action recognition. Pooling spatio-temporal descriptors with Bag of Features (BOF) technique is a widely used approach for action recognition \cite{wang2009evaluation} \cite{ma2013action}. Recently, improved Dense Trajectories features with Fisher vector encoding has been the main paradigm for local feature based video representation \cite{iDT}. In this paradigm, local descriptors are aligned and pooled along the trajectories with high motion salience and then encoded by effective Fisher vector. However, temporal information is not well utilized in deep video representation and thus constraints the performance improvement.

For deep video representations \cite{Thumos14-1}, video frames are regarded as still images and inputs of the trained ConvNets to extract features of the fully-connected layer. Average pooling across frames are then used to get the video features. As an improvement, Xu et al. \cite{LCD} use multi-scale pooling on the pooling$_5$ layer to get latent concept descriptors and encode them by VLAD. However, in above methods, no temporal variations are used in the pooling and coding phase. In 3D-CNN \cite{3D-CNN} and Deep ConvNets \cite{DeepConvNets}, they directly used the 3D video cubes and modified the image ConvNets for video classification. Though motion information can be embedded in video cubes, it takes large computational costs and the improvements are not significant. The most successful architecture which competes the state-of-the-art performance of improved Dense Trajectories is the Two-Stream ConvNets \cite{two-stream}. It is composed of two neural networks (namely spatial nets and temporal nets) aiming to capture the discriminative appearance features and motion features in one framework. Unlike deep image ConvNets which overwhelmed the other feature engineering methods, deep video representation needs to be well improved. How to properly utilize the intrinsic characteristics of videos to pool and encode should be important and essential.

Motivated by above discussions, we propose an efficient video representation framework. We get the benefits from two state-of-the-art ConvNets: VGGNet \cite{VGG} and temporal nets from Two-Stream ConvNets \cite{two-stream}. In our framework, Trajectory pooling and line pooling are used together to pool the extracted convolutional layers and the new proposed frame-diff layers to get local descriptors. We then use VLAD to encode the pooled local descriptors and form the final representations.

We illustrate our framework in Figure \ref{fig:big_pic}. For spatial ConvNets, we extract the convolutional layers and frame-diff layers from the trained VGGNet. The frame-diff layers are generated from original convolutional layers and the goal is to capture the motion information in consecutive frames. For temporal ConvNets, we extract convolutional layers from the optical flow nets of Two-Stream ConvNets. We use detected trajectories points from improved Dense Trajectories and line points to locate certain interesting points on convolutional feature maps. The responses from feature maps of one frame are stacked across the channel to form descriptors. Then the descriptors from the same line or trajectory are average pooled. At last, we choose the VLAD encoding strategy to aggregate these local descriptors for final video features and use multi-class linear SVM for action recognition. We conduct experiments on two public action datasets: HMDB51\cite{HMDB51} and UCF101 \cite{UCF101}. We get the state-of-the-art results on UCF101 and comparable to the state-of-the-art on HMDB51.

The rest of this paper is organized as follows. In Section 2, we briefly review the hand-crafted features and deep learning methods. In section 3, we propose the convolutional layer extraction and pooling framework in detail. Encoding strategy is described in Section 4. Then we report the experimental results and discussions in Section 5. Finally, we conclude our paper in Section 6.

\section{Related work}
Analogy to image classification, early researches of action recognition widely used local descriptors with BOF model such as 3D Histogram of Gradient (HOG3D) \cite{HOG3D}, and Extended SURF (ESURF) \cite{ESURF}. The difference between images is that these local descriptors are extracted and pooled over sparse spatial-temporal interesting points. In \cite{stip}, Harris3D detector is used to detect the informative regions and the interesting points are described by Histogram of Gradient (HOG) and Histogram of Optical Flow (HOF) \cite{stip}. In \cite{mosift}, Sift key points and corresponding optical flows of the same scale are detected and extracted. Then they are described by HOG and HOF respectively. Instead of computing local features over spatial-temporal cuboids, the state-of-the-art local features (i.e., iDT) \cite{iDT} detects the dense point trajectories and then pools local features along the trajectories to form local descriptors with HOG, HOF and Motion Boundary Histogram (MBH). Fisher vector is then used to aggregate these local descriptors over the whole video into a global super vector.

Dense Trajectories and its improved version have dominated action recognition for a period of time, owing to their rich captured spatial-temporal information. However, it suffers from the problem of huge computation costs \cite{LCD} and large disk affords \cite{sapienza2014feature}. For large-scale video tasks like Thumos Challenge \cite{jiang2014thumos} and TrecvId \cite{TrecvId}, it is not the best choice to use iDT due to the efficiency problem.

Inspired by the great success in deep image classification, a series of attempts have been made for video action recognition \cite{two-stream} \cite{TDD} \cite{LCD} \cite{3D-CNN} \cite{DeepConvNets}. In \cite{Thumos14-1}, video frames are regarded as still images to extract fully-connected layer features. Then average pooling is made across frames to get video features. Xu et al. \cite{LCD} used multi-scale pooling on the pooling$_5$ layer to get latent concept descriptors and encoded them by VLAD for event detection. However, temporal motion information is not employed in these methods. In order to learn the motion features, Ji et al. \cite{3D-CNN} changed the first convolutional layer to extend 2D ConvNets to videos for action recognition on relatively small datasets.  Karpathy et al. \cite{DeepConvNets} used different time fusion strategies and trained the ConvNets on a large dataset, called Sports-1M. Recently, Simonyan et al. \cite{two-stream} designed Two-Stream ConvNets containing spatial and temporal nets aiming to capture the discriminative appearance feature and motion feature, which competes the state-of-the-art performance.

However, unlike the overwhelmed advantages over traditional representation methods in images, deep video representation needs to be well improved. How to properly utilize the intrinsic characteristics of videos to incorporate motion and appearance information should be important. Particularly, network and layer selection, pooling and encoding strategy are all the important issues. In this paper, we propose a novel framework which can effectively pool the convolutional lays of spatial and temporal ConvNets. Experimental results demonstrate that our method can get the state-of-the-art performance of action recognition.

\section{Pooling the convolutional layers}
In this section, we describe our framework of pooling the convolutional layers for action recognition. We first introduce the spatial and temporal deep ConvNets used in our framework. Then we show how to extract feature maps from spatial convolutional layers, frame-diff layers and temporal convolutional layers. We also describe two pooling strategies: Trajectory pooling and line pooling.
\subsection{Spatial and temporal deep ConvNets}
\textbf{Spatial ConvNets.} As described above, video actions are more complex compared with still images and need more powerful and effective models. Thus we take the very deep convolutional network VGGNet \cite{VGG} as the spatial nets, which is the winner in ImageNet Challenge 2014. It is composed of 8 parts: 5 groups of convolutional layers and 3 fully-connected layers, which is similar to the architecture of AlexNet \cite{AlexNet}. The difference is that it has smaller convolutional size ($3\times3$), smaller convolutional stride ($1\times1$), smaller pooling window ($2\times2$) and deeper structure (up to 19 layers). Two variants, VGG-16 and VGG-19, are both successful models for ImageNet challenge. In our framework, we take VGG-16 as spatial nets.

\textbf{Temporal ConvNets.} Directly using spatial network or modifying it to 3D ConvNets cannot obtain ideal performance, because of the lack of motion information \cite{DeepConvNets} \cite{3D-CNN}. Two-Stream ConvNets is composed of two separate ConvNets, namely spatial nets and temporal nets. They are designed for capturing static appearance cues and dynamic motion information respectively. The spatial nets are trained on single frames ($224\times224\times3$), while the temporal nets are trained on stacked optical flow fields ($224\times224\times2F$, $F$ is the number of stacked flows). The spatial nets benefits from pre-training on ImageNet and temporal nets contains motion information with optical flow. Though Two-Stream ConvNets matches the state-of-the-art performance of improved Dense Trajectories, its spatial nets are not the ConvNets with the best performance. In our framework, we only choose the temporal nets from Two-Stream ConvNets.

\subsection{Extracting the convolutional layers}
As we all know, in image ConvNets, different layers of deep networks can express different information. The fully-connected layer usually denotes high-level concepts. And deeper convolutional layers contain global expressions such as object and scene, while shallower convolutional layers contain local characteristics of the image like lines, edges. At first, fully-connected layers are applied in image and video classifications and achieve good performance \cite{DeCAF} \cite{Thumos14-1}. Recently, other layers like pooling layer \cite{LCD}, convolutional layers \cite{liu2014treasure} \cite{ng2015exploiting} are also extracted and utilized. We propose to extract the convolutional layers and frame-diff layers in our framework.

Before introducing the extracted layers, we introduce some notations at first. $x$, $y$ and $t$ denote the horizontal, vertical and temporal positions of the action videos. $u$ and $v$ denotes the horizontal and vertical positions in a specific layer's feature maps. $C$ is the convolutional feature map and $D$ is the proposed frame-diff layer feature maps. $C_l^t$ denotes the feature maps from $l$-th convolutional layer which is extracted on frame $t$. $C_l^t(u, v)$ is a vector by concatenating all feature maps at position $(u, v)$.

As a very deep ConvNets, VGGNet contains 5 groups of convolutional layers. We choose the last convolutional layer in the last group and denote it as conv$_5$. For temporal ConvNets, temporal nets of Two-Stream ConvNets contains 5 convolutional layers and we choose the third and fifth convolutional layers denoted as conv$_3$ and conv$_5$. These three convolutional layers have the same size of $14\times 14$ and contain rich spatial and temporal information.

To extract and incorporate motion information in consecutive frames from spatial ConvNets, we design the frame-diff layer diff$_5$ from original spatial convolutional layers conv$_5$. As shown in Figure \ref{fig:big_pic}, when we get $C_l^t$ in frame $t$ from layer $l$ and $C_l^{t+1}$ from frame $t+1$, the frame-diff layer is as follows:
\begin{align}
        D_l^t(u, v) = C_l^{t+1}(u, v) - C_l^{t}(u, v).
        \label{eq:frame_diff}
\end{align}
Spatial-temporal normalization \cite{TDD} are then applied to the extracted convolutional layers and frame-diff layer across the video.

\subsection{Pooling the convolutional feature maps}
Once we have the convolutional feature maps, two pooling strategies are employed to get video descriptors.

\textbf{Trajectory Pooling.} This pooling strategy is based on the Dense Trajectories \cite{DT} and improved Dense Trajectories \cite{iDT}. To extract dense trajectories, feature points are sampled from multiple spatial scales. Each point $P_t=(x_t, y_t)$ at frame $t$ is tracked to the next frame $t+1$ by median filtering in a dense optical flow field $\omega = (u_t, v_t)$.
\begin{align}
        P_{t+1} = (x_{t+1}, y_{t+1}) = (x_t, y_t) +(M*\omega)|_{\hat{x}_t, \hat{y}_t},
        \label{eq:DT}
\end{align}
where $*$ is convolutional operation, $M$ is the median filter kernel, and $(\hat{x}_t, \hat{y}_t)$ is the rounded position of $(x_t, y_t)$. Once dense optical flow field is computed, points of subsequent frames are tracked and concatenated to form a trajectory: $(P_t, P_{t+1}, P_{t+2}, \dots)$. To avoid the drifting problem of tracking, the maximum length of trajectory is set to 15 frames. As an improvement, improved Dense Trajectories cancels out camera motion from optical flow to boost the performance. It assumes that global background motion of two consecutive frames are related by a homography matrix. In order to estimate the homography, the correspondences between two frames are found by two complementary matching strategies: SURF \cite{surf} feature matching and optical flow matching. RANSAC \cite{RANSAC} algorithm is used to robustly estimate the homography which allows to rectify the image to remove the camera motion. In addition, when estimating the homography, it uses the human detector as a mask to remove feature matches inside the bounding boxes.

An extracted trajectory can be denoted as:
\begin{align}
        T^k=\{(x_1^k, y_1^k, t_1^k), \dots, (x_N^k, y_N^k, t_N^k)\}.
        \label{eq:Traj}
\end{align}
Here, $T^k$ is the $k$-th trajectory with $N$ (15 in our framework) points, totally there are $K$ trajectories. In order to pool the convolutional maps using trajectories, we need to compute a mapping ratio which maps the video points to feature maps. The ratio $\alpha$ is computed as:
 \begin{align}
        \alpha = \frac{H_f^l}{H_v} = \frac{W_f^l}{W_v}.
        \label{eq:ratio}
\end{align}
Here, $H_f^l$ and $W_f^l$ denote the height and width of the $l$-th convolutional feature maps, while $H_v$ and $W_v$ denote the height and width of video frames.
The $i$-th trajectory descriptor $TD_i^l$ from $l$-th layer can be computed as:
 \begin{align}
        TD_i^l = \sum\limits_{i=1}^N  C_{l}^{t_i^k} (x_i^k\times \alpha, y_i^k\times \alpha)
        \label{eq:TD}
\end{align}

\textbf{Line Pooling.}
Despite the good performance, trajectory pooling suffers from the efficiency problem. So we propose an alternative pooling strategy, called line pooling. Instead of pooling along the time-consuming trajectories which are computed from original video, line pooling directly pools stacked features from the convolutional feature maps along the time line. Specifically, a line is denoted as:
\begin{align}
        L^k=\{(u^k, v^k, t_1^k), \dots, (u^k, v^k, t_N^k)\}.
        \label{eq:Line}
\end{align}
where $u^k \in [1, H_f^l]$ and $v_k \in [1, W_f^l]$.
With this pooling line, the $i$-th line descriptor $LD_i^l$ from $l$-th layer can be computed as:
 \begin{align}
        LD_i^l = \sum\limits_{i=1}^N  C_{l}^{t_i^k} (u^k, v^k).
        \label{eq:LD}
\end{align}

Compared with Trajectory pooling, the line pooling has the following characteristics:
\begin{itemize}
    \item It pools convolutional features directly on feature maps along the time line, thus it can skip the pre-computing on video actions and thus is faster.
    \item The number of pooled descriptors of a video (containing $T$ frames) is $H_f^l \times W_f^l \times T$ by line pooling, which is relatively fixed. While in trajectory pooling, the number depends on trajectories numbers $K$ and is uncertain.
    \item Line pooling pools all the feature points in convolutional layers and may contain noise and useless information, while trajectory pooling can make use of the dense trajectories extracted from video sequences.
\end{itemize}

\section{Local feature encoding}
\subsection{Fisher vector Encoding}
The Fisher vector encoding is based on a fitted parametric generative model, e.g. the Gaussian Mixture Model (GMM) and it encodes the local features with respect to the first-order and second-order parameters. A Gaussian Mixture Model (GMM) with $K$ components can be denoted as $\Theta = \{(\mu_k, \pi_k, \sigma_k)\}, k=1, 2, \dots, K$, where $\mu_k$, $\Sigma_k$, $\pi_k$ are the mixture weights, means, and diagonal covariances of the GMM, which are computed on the pooled descriptors of the training set. $\alpha_k(x_p)$ is the soft assignment weight of the $p$-th descriptor to $k$-th Gaussian. Given $X=(x_1, x_2,\dots, x_N)$ extracted and pooled from a video, a Fisher vector component is computed as:

\begin{align}
        \nonumber
        \Phi_k^{(1)} = \frac{1}{N\sqrt{\pi_k}} \sum\limits_{p=1}^N \alpha_k(x_p) (\frac{x_p-\mu_k}{\sigma_k}),
\end{align}
\begin{align}
        \Phi_k^{(2)} = \frac{1}{N\sqrt{2\pi_k}} \sum\limits_{p=1}^N \alpha_k(x_p)
        \big(\frac{(x_p-\mu_k)^2}{\sigma_k^2}-1\big).
        \label{eq:FV}
\end{align}

Then the Fisher vector representation is obtained by stacking these components: $\phi = [\Phi_1^{(1)}, \Phi_1^{(2)}, \dots, \Phi_K^{(1)}, \Phi_K^{(2)}]$. Each video is finally represented by a $2D'K$ dimension vector, where $D'$ is the dimension of pooled descriptor $x_i$ after PCA pre-processing. PCA is used to decorrelate the descriptors and better fit on the diagonal covariance matrix assumption. Usually, power normalization with $z$=sign$(z)\sqrt{|z|}$ and $\ell_2$ normalization are applied to Fisher vector.
\subsection{VLAD encoding}
VLAD \cite{VLAD} aggregates descriptors based on a locality criterion in feature space. It can be seen as a simplified version of the Fisher vector. Similar to BOF \cite{BOF}, it first learns a codebook $\mathcal{C} = \{c_1, \dots, c_K\}$ of $K$ visual words with $k$-means. Each local descriptor is assigned according to its nearest visual word $c_i = NN(x)$, $NN(x)$ denotes $x$'s nearest neighbor. The idea of VLAD is to accumulate the differences $x-c_i$ of vectors $x$ and the certain center $c_i$. This can represent the characteristic of vector distribution over the generated centers.

Given pooled descriptors of $X=(x_1, x_2, \dots, x_N)$, the difference vector computed from center $k$ is:
 \begin{align}
        \mu_k = \sum\limits_{i:NN(x_i)=c_k}(x_i-c_k),
        \label{eq:VLAD}
\end{align}

Then VLAD representation is obtained by concatenating $\mu_k$ over all the $K$ centers as: $\mu = [\mu_1, \mu_2, \dots, \mu_K]$. Power and $\ell_2$ normalization are also used to post-process on the representation. In addition, we apply intra-normalization which adds normalization on each $u_k$. In our framework, we use a variant of VLAD called VLAD-$k$ ($k=5$), which extends nearest neighbor to $k$-nearest neighbors and has shown good performance in action recognition \cite{peng2014bag} \cite{kantorov2014efficient}.

Fisher vector and VLAD are all efficient and effective encoding methods for local descriptors. Fisher vector is the default encoding for improved Dense Trajectories and VLAD is applied to deep latent concept descriptors in \cite{LCD}. Since VLAD encoding is simpler and generates lower dimension features with the same centers $K$, we choose it to encode the pooled convolutional feature maps.

\section{Experiments}
In this section, we first describe the datasets used for evaluating the performance. Then, we conduct exploration experiments to determine some crucial factors. In the following, we evaluate the networks, layers and pooling strategies successively. At last, comparison to the state-of-the-art methods are presented.
\subsection{Datasets}
In our experiments, we choose two widely-used large-scale public datasets: HMDB51 \cite{HMDB51} and UCF101 \cite{UCF101}. HMDB51 dataset is a real-world dataset containing complicated videos collected from movies and web videos. This dataset includes 6,766 video clips from 51 action classes with each class containing at least 100 video clips. It is divided into three different splits. For each split, there are 70 video clips used for training and 30 clips used for testing in each action class.	

The UCF101 dataset contains 101 action classes with each class including at least 100 video clips. It is divided into 25 groups according to the action performer. There are 13,320 trimmed video clips in this dataset. The same as HMDB51 dataset, UCF101 is divided into three training/testing splits for performance evaluation.

For these two datasets, we follow the default training/test splits of each dataset and report the average accuracy of the three splits.

\subsection{Exploration experiments}

\begin{table}
\begin{center}
\begin{tabular}{|l|l|c|}
\hline
 Dataset & VLAD-$k$ & Fisher vector \cite{TDD}\\
\hline\hline
UCF101 & 82.90\% & 81.7\%  \\
\hline
HMDB51 & 55.73\% & 54.5\%\\
\hline
\end{tabular}
\end{center}
\caption{Performance of different encoding strategy on HMDB51 and UCF101 dataset. VLAD-$k$ shows better performance.}
\label{tab:encode}
\end{table}

\begin{figure}[ht]

\begin{minipage}[b]{.42\linewidth}
  \centering
  \centerline{\epsfig{figure=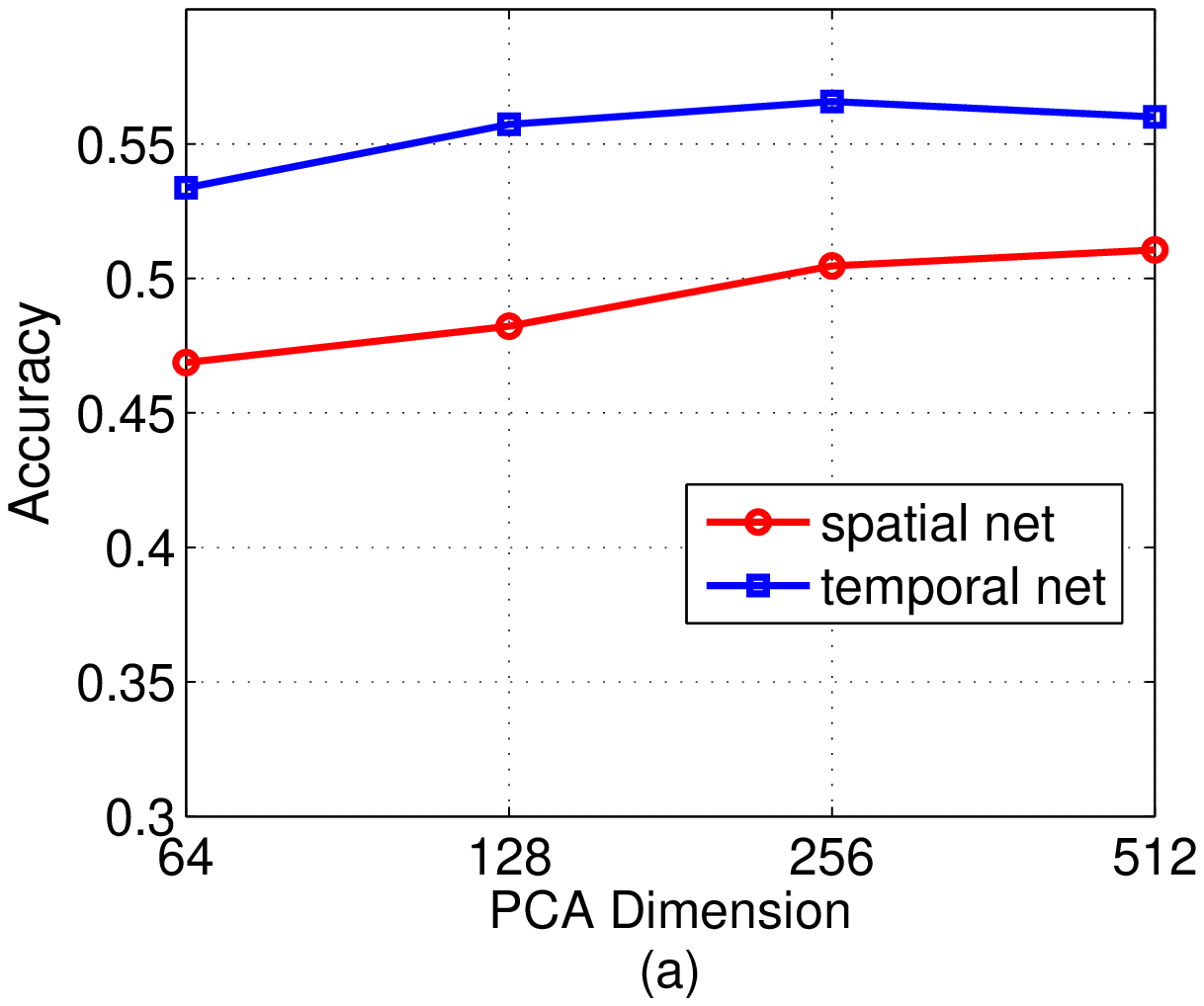,width=4.0cm}}
\end{minipage}
\hfill
\begin{minipage}[b]{0.42\linewidth}
  \centering
  \centerline{\epsfig{figure=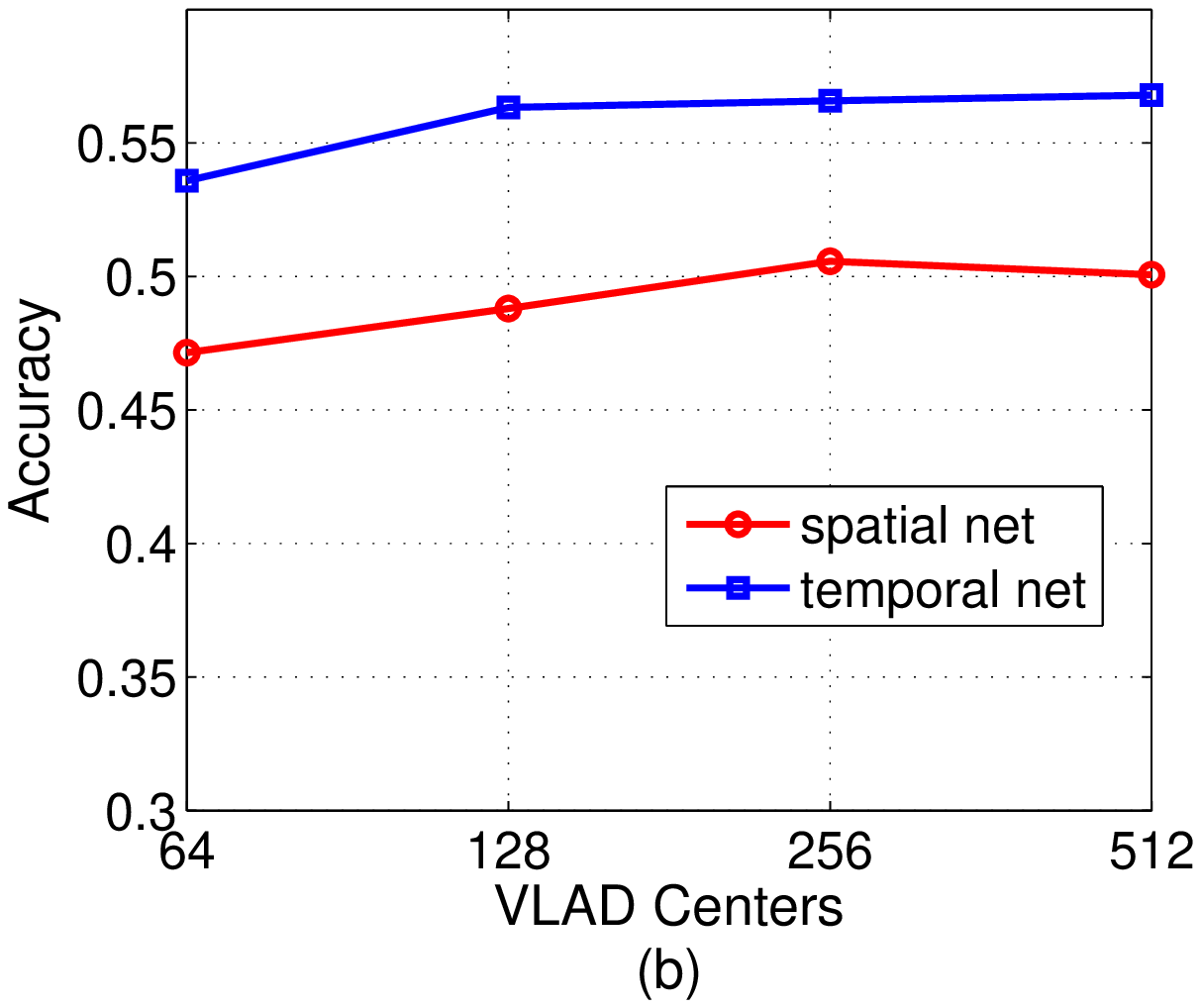,width=4.0cm}}
\end{minipage}

\caption{Exploration of PCA dimension (a) and VLAD centers (b) on spatial net and temporal net in HMDB51 dataset.}
\label{fig:PCA_Center}
\end{figure}

\textbf{Dimension of PCA.} In local descriptors like improved Dense Trajectories and deep descriptors like LCD \cite{LCD}, PCA with whitening is usually applied to de-correlate the descriptors and reduce the dimension. The dimension of pooled descriptors is 512. We evaluate the performance with reduced dimensions varying from 64 to 512 on trajectory pooled spatial conv$_3$ and temporal conv$_3$ layers on HMDB51 dataset. Results are shown in Figure \ref{fig:PCA_Center} (a). To balance the accuracy and feature dimension, we choose 256 in the rest of our experiments.

\textbf{Feature encoding.} Fisher vector and VLAD are popular encoding strategies. To evaluate the performances, we extracted temporal conv$_3$ layer with Trajectory pooling and compare it with \cite{TDD}. To ensure the same feature dimension, we set $K=128$ in VLAD-$k$ ($k$=5) as $K=64$ is set in Fisher vector in \cite{TDD}. Results in Table \ref{tab:encode} shows that VLAD-$k$ has better performance than Fisher vector. So we utilize VLAD-$k$ as the default encoding strategy in our experiments.

\textbf{VLAD centers.} In VLAD encoding, number of centers $K$ determines the final feature dimension. We perform trajectory pooling on spatial conv$_3$ and temporal conv$_3$ layers in HMDB51 dataset to explore the best $K$. As shown in Figure \ref{fig:PCA_Center} (b), the discriminative ability of the generated features improves when $K$ increases until 256. As a result, we fix the number of VLAD centers to 256 in our experiments to obtain good performances.

\textbf{Action classifier.} In action recognition, the generated feature dimension is usually very high. So we apply one-versus-rest multi-class linear Support Vector Machine (SVM) as the classifier. The libSVM \cite{libSVM} implementation is adopted in our experiments. As to the parameter $C$, we follow the common setting as in \cite{LCD} \cite{Thumos14-1} to set it to 100.

\subsection{Evaluation of networks}
In temporal ConvNets, we utilize the temporal nets of Two-Stream ConvNets which accepts stacked optical flow fields as input. As till now, the temporal nets of Two-Stream ConvNets is the state-of-the-art temporal network for action recognition. And because network training is not the key point in this paper, we skip the fine-tune process of the networks and pay more attention to our strategy. In our experiments, we choose the model trained by \cite{TDD} as our temporal ConvNets. We will show that better temporal networks can further improve our performance.

Though temporal nets in Two-Stream ConvNets have shown good performance, the spatial nets are not the state-of-the-art. For spatial ConvNets, VGGNet \cite{VGG} is a winner of ImageNet Challenge 2014. To evaluate the spatial networks, we utilize Trajectory pooling on conv$_5$ layer of different spatial networks on HMDB51 and UCF101 datasets. Table \ref{network} shows that VGGNet outperforms spatial nets of Two-Stream ConvNets by 4\% on HMDB51 and 5\% on UCF101. We analyze this results from two aspects: on one hand, VGGNet is much deeper than spatial nets of Two-Stream ConvNets, which can model complex action concepts and scenes; on the other hand, VGGNet is trained on ImageNet dataset, which is larger than UCF101 and HMDB51. So in our framework, we use the pre-trained VGGNet to replace the spatial nets of Two-Stream ConvNets and generate our deep spatial-temporal ConvNets.

\begin{table}[t]
\begin{center}
\begin{tabular}{|l|l|l|}
\hline
Network                      & UCF101 & HMDB51  \\
\hline\hline
Two-Stream Spatial \cite{TDD} & 79.79\% & 45.51\% \\
\hline
VGGNet                       & 84.97\% & 50.42\% \\
\hline
\end{tabular}
\end{center}
\caption{The performance comparison of different networks. Trajectory pooling is applied on the conv$_5$ layer of VGG net and spatial network of Two-Stream ConvNets on UCF101 and HMDB51.}
\label{network}
\end{table}

\subsection{Evaluation of different layers}
In this subsection, we investigate the performance of different layers in deep ConvNets, including the fully-connected layers, convolutional layers and proposed frame-diff layer. Trajectory pooling is applied on convolutional layers while average pooling is applied on fully-connected layers. We start with temporal networks on HMDB51 dataset to evaluate different layers and layer combinations. The results are presented in Table \ref{layers}.

\begin{table}[h]
\begin{center}
\begin{tabular}{|l|l|}
\hline
           Layer  & HMDB51   \\
\hline\hline
Fc$_6$            & 45.10\% \\
Fc$_7$            & 41.72\% \\
Fc$_6$+Fc$_7$     & 45.25\% \\
\hline
Conv$_3$          & 56.58\% \\
Conv$_4$          & 57.03\% \\
Conv$_5$          & 51.92\% \\
\hline
Conv$_3$+Conv$_4$ & 58.58\% \\
Conv$_4$+Conv$_5$ & 57.82\% \\
Conv$_3$+Conv$_5$ & 59.24\% \\
\hline
Conv$_3$+Conv$_4$+Conv$_5$ & 59.91\% \\
\hline
\end{tabular}
\end{center}
\caption{Comparison of different temporal layers and layer combinations on HMDB51 dataset. We investigate the fully-connected layers and convolutional layer combinations. We conduct combination by early fusion \cite{oneata2014lear}.}
\label{layers}
\end{table}

The fully-connected layers perform not so well compared with convolutional layers. This is because the fully-connected layers are feature vectors and lack local and regional information. As to single convolutional layer, Conv$_4$ achieves the best performance while Conv$_3$+Conv$_5$ achives the best among combinations of two layers. Combination of three layers improves 0.67\% compared to the best combination of two layers. But the combination of three layers will lead to higher dimension and larger time costs, which may not be a good choice.

We further investigate the best combination Conv$_3$+Conv$_5$ on UCF101 and HMDB51 datasets. From Table \ref{Conv35}, we can see that Conv$_3$ is more suitable for complex and noisy videos in HMDB51, while Conv$_5$ applies to relatively trimmed videos in UCF101. The combination performance has 0.93\% decrease on UCF101 and 2.66\% improvement on HMDB51. To get a general video representation, we combine temporal Conv$_3$ and Conv$_5$ in our framework.

\begin{table}
\begin{center}
\begin{tabular}{|l|l|l|}
\hline
Layer             & UCF101 & HMDB51   \\
\hline\hline
Conv$_3$          & 83.99\% & 56.58\% \\
Conv$_5$          & 90.18\% & 51.92\% \\
\hline
Conv$_3$+Conv$_5$ & 89.11\% & 59.24\% \\
\hline
\end{tabular}
\end{center}
\caption{The performance of temporal Conv$_3$, Conv$_5$ and its combination on UCF101 and HMDB51 datasets.}
\label{Conv35}
\end{table}

Compared with temporal network, the spatial VGGNet is deeper and more powerful. As there are 16 layers, it is difficult to evaluate each layer separately. To align with temporal network in feature map size, we directly choose Conv$_5$ layer of VGGNet as the convolutional layer. As VGGNet only contains spatial information of video frames, we propose to extract motion information of consecutive frames via a small modification: using the frame-diff layer. This is actually similar as the optical flow but saving the time of training a new network.

\begin{table}
\begin{center}
\begin{tabular}{|l|l|l|}
\hline
Layer             & UCF101  & HMDB51   \\
\hline\hline
Conv$_5$          & 84.97\% & 50.57\% \\
Diff$_5$          & 83.94\% & 50.00\% \\
\hline
Conv$_5$+Diff$_5$ & 86.53\% & 53.27\% \\
\hline
\end{tabular}
\end{center}
\caption{The performance of spatial Conv$_5$ and Diff$_5$ on UCF101 and HMDB51 datasets. The combination of Diff$_5$ and Conv$_5$ can improve the performance in a degree.}
\label{Diff5}
\end{table}

In Tabel \ref{Diff5}, the performance of Diff$_5$ layer is slightly worse than Conv$_5$. However, as is shown in Table \ref{Diff5}, the incorporation of Diff$_5$ can boost the performance of Conv$_5$ by 1.56\% on UCF101 and 2.70\% on HMDB51. We can also consider that frame-diff layer is a coarse approximation of the optical flow layer. So, in some scenarios, when the optical flow network is difficult to train, we can have a powerful spatial network at hand. Thus, it is a good choice to extract motion information using our proposed frame-diff layer.

\begin{figure}[ht]
\begin{center}
\epsfig{file=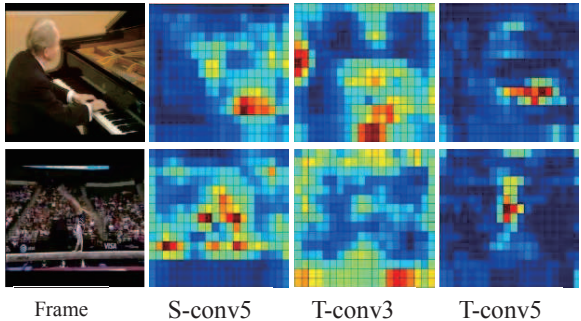, width=0.45\textwidth}
\end{center}
   \caption{Convolutional layer visualization of UCF101. S-conv5 denotes spatial Conv$_5$ layer and T-conv3, T-conv5 denote temporal Conv$_3$ and Conv$_5$ respectively.}
\label{fig:visual}
\end{figure}

\textbf{Layer visualization}. The extracted convolutional layers of action ``Playing Piano" and ``Balance Beam" on UCF101 are shown in Figure \ref{fig:visual}. We can see that spatial Conv$_5$ corresponds well to action scenes. Moreover, Conv$_5$ is more sparse and more accurate than Conv$_3$ with respect to action motions. Thus the temporal Conv$_5$ is proper for trimmed datasets like UCF101.

\subsection{Evaluation of pooling strategies}

Based on the above experiments and discussions, we pool the Conv$_3$ and Conv$_5$ layers of temporal network and pool the Conv$_5$ and Diff$_5$ layers of spatial network.

The performance comparison of three pooling strategies to the original Two-Stream ConvNets is presented in Table \ref{Pooling}. TDD \cite{TDD} uses the same pooling strategy as ours but it utilizes different networks and encoding strategies. From the results we can see that all three pooling strategies outperform the original Two-Stream ConvNets. The results successfully demonstrate that the effective pooling strategy can better exploit the intrinsic characteristic of videos. Moreover, from Table \ref{Pooling}, our line pooling strategy is comparable to TDD and the Trajectory pooling has more than 2\% improvement on TDD. This shows the effectiveness of our proposed framework which successfully benefits from the better network architecture and layer design. Though line pooling is not as good as Trajectory pooling because of the lack of prior trajectory knowledge, line pooling can also be used as a fast version of video representation as it can gets comparable performance.
 	
\begin{table}
\begin{center}
\begin{tabular}{|l|l|l|}
\hline
                   & UCF101 & HMDB51   \\
\hline\hline
Two Stream \cite{two-stream}   & 88.0\% & 59.4\%  \\
TDD \cite{TDD}          & 90.3\%& 63.2\% \\
Line Pooling       & 90.55\%& 62.24\% \\
Trajectory Pooling & \textbf{93.78\%}& \textbf{65.62\%} \\
\hline
\end{tabular}
\end{center}
\caption{The performance comparison of three pooling strategies with the original Two-Stream ConvNets. TDD \cite{TDD} uses the same Trajectory pooling as ours but adopts different networks and encoding methods.}
\label{Pooling}
\end{table}

\subsection{Comparison to the state-of-the-art}

Table \ref{State} compares our recognition results with several recently published methods on the dataset of HMDB51 and
UCF101. These methods can be divided into three types: (1) hand-crafted local features like iDT, (2) deep Two-Stream ConvNets and its variants, (3) pooling on deep ConvNets like TDD and our method.

Compared with hand-crafted features, our line pooling method has a 2\% advantage on UCF101 and 1\% on HMDB51, while Trajectory pooling has a 5\% and 4\% advantages. These hand-crafted features rely on human knowledge to design and optimize. Our deep learned features benefit from both the large image datasets from which we can get the appearance information, and the motion information obtained from the optical flow net.

\begin{table}
\begin{center}
\begin{tabular}{|l|l|l|}
\hline
Algorithm                              & HMDB51 & UCF101 \\
\hline\hline
DT+BoVW \cite{DT}                      & 46.6\% & -      \\
DT+VLAD \cite{cai2014multi}            &  -     & 79.9\% \\
DT+MVSV \cite{cai2014multi}            & 55.9\% & 83.5\% \\
iDT+FV \cite{UCFiDT+FV}                & 57.2\% & 85.9\% \\
iDT+HSV \cite{iDT+HSV}                 & 61.1\% & 87.9\% \\
\hline
Two-Stream \cite{two-stream}           & 59.4\% & 88.0\% \\
Two-stream+LSTM \cite{Two-stream+LSTM} & -      & 88.6\% \\
Very deep two-stream \cite{Very-deep}  & -      & 91.4\% \\
\hline
TDD \cite{TDD}                         & 63.2\% & 90.3\% \\
TDD+iDT \cite{TDD}                     & \textbf{65.9\%} & 91.5\% \\
Line Pooling                           & 62.24\% & 90.55\% \\
Trajectory Pooling                     & 65.62\% & \textbf{93.78\%} \\
\hline
\end{tabular}
\end{center}
\caption{Comparison to the state-of-the-art methods.}
\label{State}
\end{table}

Two-Stream ConvNets  use spatial and temporal networks to capture both the appearance and motion information and obtain comparable performance with traditional improved Trajectories. \cite{Two-stream+LSTM} adds LSTM model to better model the video sequences. Note that a very recent proposed model in \cite{Very-deep} utilizes and trains on very deep ConvNets and achieves 91.4\% on UCF101. From the results in Table \ref{State} we can see, our method outperforms these methods on two aspects: network design and pooling strategy. With different convolutional layers and the generated frame-diff layer, complementary spatial-temporal information of deep networks is successfully utilized. Moreover, temporal action scene and motion are effectively pooled and encoded by our pooling and encoding strategy.

At last, we compare our method with other pooling approaches. From the results we can see, line pooling is comparable to TDD \cite{TDD} and Trajectory pooling has more than 2\% advantage over TDD. This improvement owes to the architecture of our framework that introduces the frame-diff layer and also the utilization of deeper networks with proper layer combinations. Compared to the combination of TDD and iDT, our method has more than 2\% advantage on UCF101 and comparable performance (i.e., only a gap less than 0.3\%) on HMDB51.

\section{Conclusion}
In this paper, we propose to pool the convolutional layers in deep ConvNets for action recognition. The network architecture is designed on two state-of-the-art deep ConvNets as the spatial and temporal networks. We extract the convolutional layers and frame-diff layer and then pool them by two strategies: Trajectory pooling and line pooling. VLAD-$k$ is employed as the encoding approach. Our method achieves the state-of-the-art accuracy on UCF101 and comparable to the state-of-the-art accuracy on HMDB51. In the future, we will train and utilize deeper temporal network to handle complex video concepts and explore more effective pooling strategies on different layers.


{\small
\bibliographystyle{ieee}
\bibliography{egbib}

\begin{thebibliography}{10}\itemsep=-1pt

\bibitem{TrecvId}
Trecvid.
\newblock \url{http://trecvid.nist.gov/}.

\bibitem{human-action1}
J.~K. Aggarwal and M.~S. Ryoo.
\newblock Human activity analysis: A review.
\newblock {\em ACM Computing Surveys (CSUR)}, 43(3):16, 2011.

\bibitem{surf}
H.~Bay, T.~Tuytelaars, and L.~Van~Gool.
\newblock Surf: Speeded up robust features.
\newblock In {\em ECCV}. 2006.

\bibitem{cai2014multi}
Z.~Cai, L.~Wang, X.~Peng, and Y.~Qiao.
\newblock Multi-view super vector for action recognition.
\newblock In {\em CVPR}, 2014.

\bibitem{BOF}
Y.~Cao, C.~Wang, Z.~Li, L.~Zhang, and L.~Zhang.
\newblock Spatial-bag-of-features.
\newblock In {\em CVPR}, 2010.

\bibitem{libSVM}
C.-C. Chang and C.-J. Lin.
\newblock Libsvm: A library for support vector machines.
\newblock {\em ACM Transactions on Intelligent Systems and Technology (TIST)},
  2(3):27, 2011.

\bibitem{mosift}
M.-y. Chen and A.~Hauptmann.
\newblock Mosift: Recognizing human actions in surveillance videos.
\newblock 2009.

\bibitem{DeCAF}
J.~Donahue, Y.~Jia, O.~Vinyals, J.~Hoffman, N.~Zhang, E.~Tzeng, and T.~Darrell.
\newblock Decaf: A deep convolutional activation feature for generic visual
  recognition.
\newblock In {\em ICML}, 2014.

\bibitem{RANSAC}
M.~A. Fischler and R.~C. Bolles.
\newblock Random sample consensus: a paradigm for model fitting with
  applications to image analysis and automated cartography.
\newblock {\em Communications of the ACM}, 24(6):381--395, 1981.

\bibitem{Thumos14-1}
M.~Jain, J.~van Gemert, and C.~G. Snoek.
\newblock University of amsterdam at thumos challenge 2014.
\newblock {\em ECCV THUMOS Challenge}, 2014.

\bibitem{VLAD}
H.~J{\'e}gou, M.~Douze, C.~Schmid, and P.~P{\'e}rez.
\newblock Aggregating local descriptors into a compact image representation.
\newblock In {\em CVPR}, 2010.

\bibitem{3D-CNN}
S.~Ji, W.~Xu, M.~Yang, and K.~Yu.
\newblock 3d convolutional neural networks for human action recognition.
\newblock {\em TPAMI}, 35(1):221--231, 2013.

\bibitem{jiang2014thumos}
Y.~Jiang, J.~Liu, A.~Roshan~Zamir, G.~Toderici, I.~Laptev, M.~Shah, and
  R.~Sukthankar.
\newblock Thumos challenge: Action recognition with a large number of classes.
\newblock {\em http://crcv.ucf.edu/THUMOS14}, 2014.

\bibitem{kantorov2014efficient}
V.~Kantorov and I.~Laptev.
\newblock Efficient feature extraction, encoding, and classification for action
  recognition.
\newblock In {\em CVPR}, 2014.

\bibitem{DeepConvNets}
A.~Karpathy, G.~Toderici, S.~Shetty, T.~Leung, R.~Sukthankar, and L.~Fei-Fei.
\newblock Large-scale video classification with convolutional neural networks.
\newblock In {\em CVPR}, 2014.

\bibitem{HOG3D}
A.~Klaser, M.~Marsza{\l}ek, and C.~Schmid.
\newblock A spatio-temporal descriptor based on 3d-gradients.
\newblock In {\em BMVC}, 2008.

\bibitem{AlexNet}
A.~Krizhevsky, I.~Sutskever, and G.~E. Hinton.
\newblock Imagenet classification with deep convolutional neural networks.
\newblock In {\em NIPS}, 2012.

\bibitem{HMDB51}
H.~Kuehne, H.~Jhuang, E.~Garrote, T.~Poggio, and T.~Serre.
\newblock Hmdb: a large video database for human motion recognition.
\newblock In {\em ICCV}, 2011.

\bibitem{stip}
I.~Laptev, M.~Marsza{\l}ek, C.~Schmid, and B.~Rozenfeld.
\newblock Learning realistic human actions from movies.
\newblock In {\em CVPR}, 2008.

\bibitem{liu2014treasure}
L.~Liu, C.~Shen, and A.~van~den Hengel.
\newblock The treasure beneath convolutional layers: Cross-convolutional-layer
  pooling for image classification.
\newblock In {\em CVPR}, 2015.

\bibitem{ma2013action}
S.~Ma, J.~Zhang, N.~Ikizler-Cinbis, and S.~Sclaroff.
\newblock Action recognition and localization by hierarchical space-time
  segments.
\newblock In {\em ICCV}, 2013.

\bibitem{Two-stream+LSTM}
J.~Y. Ng, M.~J. Hausknecht, S.~Vijayanarasimhan, O.~Vinyals, R.~Monga, and
  G.~Toderici.
\newblock Beyond short snippets: Deep networks for video classification.
\newblock In {\em CVPR}, 2015.

\bibitem{ng2015exploiting}
J.~Y. Ng, F.~Yang, and L.~S. Davis.
\newblock Exploiting local features from deep networks for image retrieval.
\newblock In {\em CVPR}, 2015.

\bibitem{oneata2014lear}
D.~Oneata, J.~Verbeek, and C.~Schmid.
\newblock The lear submission at thumos 2014.
\newblock 2014.

\bibitem{peng2014bag}
X.~Peng, L.~Wang, X.~Wang, and Y.~Qiao.
\newblock Bag of visual words and fusion methods for action recognition:
  Comprehensive study and good practice.
\newblock {\em CoRR}, abs/1405.4506, 2014.

\bibitem{iDT+HSV}
S.~Sadanand and J.~J. Corso.
\newblock Action bank: A high-level representation of activity in video.
\newblock In {\em CVPR}, 2012.

\bibitem{sapienza2014feature}
M.~Sapienza, F.~Cuzzolin, and P.~H.~S. Torr.
\newblock Feature sampling and partitioning for visual vocabulary generation on
  large action classification datasets.
\newblock {\em CoRR}, abs/1405.7545, 2014.

\bibitem{two-stream}
K.~Simonyan and A.~Zisserman.
\newblock Two-stream convolutional networks for action recognition in videos.
\newblock In {\em NIPS}, 2014.

\bibitem{VGG}
K.~Simonyan and A.~Zisserman.
\newblock Very deep convolutional networks for large-scale image recognition.
\newblock {\em CoRR}, abs/1409.1556, 2014.

\bibitem{UCF101}
K.~Soomro, A.~R. Zamir, and M.~Shah.
\newblock {UCF101:} {A} dataset of 101 human actions classes from videos in the
  wild.
\newblock {\em CoRR}, abs/1212.0402, 2012.

\bibitem{szegedy2015going}
C.~Szegedy, W.~Liu, Y.~Jia, P.~Sermanet, S.~Reed, D.~Anguelov, D.~Erhan,
  V.~Vanhoucke, and A.~Rabinovich.
\newblock Going deeper with convolutions.
\newblock In {\em CVPR}, 2015.

\bibitem{DT}
H.~Wang, A.~Kl{\"a}ser, C.~Schmid, and C.-L. Liu.
\newblock Dense trajectories and motion boundary descriptors for action
  recognition.
\newblock {\em IJCV}, 103(1):60--79, 2013.

\bibitem{iDT}
H.~Wang and C.~Schmid.
\newblock Action recognition with improved trajectories.
\newblock In {\em ICCV}, 2013.

\bibitem{UCFiDT+FV}
H.~Wang and C.~Schmid.
\newblock Lear-inria submission for the thumos workshop.
\newblock In {\em ICCV Workshop}, 2013.

\bibitem{wang2009evaluation}
H.~Wang, M.~M. Ullah, A.~Klaser, I.~Laptev, and C.~Schmid.
\newblock Evaluation of local spatio-temporal features for action recognition.
\newblock In {\em BMVC}, 2009.

\bibitem{TDD}
L.~Wang, Y.~Qiao, and X.~Tang.
\newblock Action recognition with trajectory-pooled deep-convolutional
  descriptors.
\newblock In {\em CVPR}, 2015.

\bibitem{Very-deep}
L.~Wang, Y.~Xiong, Z.~Wang, and Y.~Qiao.
\newblock Towards good practices for very deep two-stream convnets.
\newblock {\em CoRR}, abs/1507.02159, 2015.

\bibitem{ESURF}
G.~Willems, T.~Tuytelaars, and L.~Van~Gool.
\newblock An efficient dense and scale-invariant spatio-temporal interest point
  detector.
\newblock In {\em ECCV}. 2008.

\bibitem{LCD}
Z.~Xu, Y.~Yang, and A.~G. Hauptmann.
\newblock A discriminative {CNN} video representation for event detection.
\newblock In {\em CVPR}, 2015.

\end{thebibliography}
}

\end{document}